\documentclass[11pt]{article}

\usepackage{colacl}

\title{Improving Data Driven Wordclass Tagging\\by System Combination}

\author{Hans van Halteren\\
Dept. of Language and Speech\\
University of Nijmegen \\
P.O. Box 9103\\
6500 HD Nijmegen\\
The Netherlands\\
hvh@let.kun.nl\\
\And
Jakub Zavrel, Walter Daelemans\\
Dept. of Computational Linguistics\\
Tilburg University\\
P.O. Box 90153\\
5000 LE Tilburg\\
The Netherlands\\
Jakub.Zavrel@kub.nl,Walter.Daelemans@kub.nl\\
}

\begin{document}
\maketitle
\bibliographystyle{acl}

\abstract{
In this paper we examine how the differences in modelling between
different data driven systems performing the same NLP task can be
exploited to yield a higher accuracy than the best individual
system. We do this by means of an experiment involving the task of
morpho-syntactic wordclass tagging. Four well-known tagger generators (Hidden
Markov Model, Memory-Based, Transformation Rules and Maximum Entropy)
are trained on the same corpus data. After comparison, their outputs are
combined using several voting strategies and second stage classifiers. 
All combination taggers outperform their best component, with the best 
combination showing a 19.1\% lower error rate than the best individual 
tagger.
}

\section*{Introduction}

In all Natural Language Processing (NLP) systems, we find one or more
language models which are used to predict, classify and/or interpret
language related observations.
Traditionally, these models were categorized as either
rule-based/symbolic or corpus-based/probabilistic. Recent work
(e.g. Brill 1992) has demonstrated clearly that this
categorization is in fact a mix-up of two distinct categorization
systems: on the one hand there is the representation used for the
language model (rules, Markov model, neural net, case base, etc.) and
on the other hand the manner in which the model is constructed (hand
crafted vs. data driven).

Data driven methods appear to be the more popular. This can be explained
by the fact that, in general, hand crafting an explicit model is rather
difficult, especially since what is being modelled, natural language,
is not (yet) well-understood. When a data driven method is used, a model is
automatically learned from the implicit structure of an annotated
training corpus. This is much easier and can
quickly lead to a model which produces results with a `reasonably'
good quality.

Obviously, `reasonably good quality' is not the ultimate goal.
Unfortunately, the quality that can be reached for a given task
is limited, and not merely by the potential of the learning method used.
Other limiting factors are the power of the hard- and software
used to implement the learning method and the availability
of training material.
Because of these limitations, we find that for most
tasks we are (at any point in time) faced with a ceiling to the quality 
that can be reached with any (then) available machine learning system.
However, the fact that any given system cannot go beyond this ceiling
does not mean that machine learning as a whole is similarly limited.
A potential loophole is that
each type of learning method brings its own `inductive bias' to
the task and therefore different methods will tend to produce different errors.
In this paper, we are concerned with the question whether these
differences between models can indeed be exploited to yield a data driven
model with superior performance.

In the machine learning literature this approach is known as ensemble,
stacked, or combined classifiers. It has been shown that, when the
errors are uncorrelated to a sufficient degree, the resulting combined
classifier will often perform better than all the individual systems
(Ali and Pazzani 1996; Chan and Stolfo 1995; Tumer and Gosh 1996). The
underlying assumption is twofold. First, the combined
votes will make the system more robust to the quirks of each learner's
particular bias. Also, the use of information about each individual
method's behaviour in principle even admits the possibility to fix
collective errors.

We will execute our investigation by means of an experiment. The NLP
task used in the experiment is morpho-syntactic wordclass tagging. The
reasons for this choice are several. First of all, tagging is a widely
researched and well-understood task
(cf.~van Halteren (ed.) 1998). Second, current performance levels on
this task still leave room for improvement: `state of the art'
performance for data driven automatic wordclass taggers (tagging
English text with single tags from a low detail tagset) is 96-97\%
correctly tagged words. Finally, a number of rather different
methods are available that generate a fully functional tagging system
{}from annotated text.

\section{Component taggers}

In 1992, van Halteren combined a number of
taggers by way of a straightforward majority vote (cf. van Halteren 1996). 
Since the component
taggers all used n-gram statistics to model context probabilities and
the knowledge representation was hence fundamentally the same in each
component, the results were limited. Now there are more varied systems
available, a variety which we hope will lead to better combination
effects.  For this experiment we have selected four systems, primarily on
the basis of availability. Each of these uses different features of
the text to be tagged, and each has a completely different
representation of the language model.

The first and oldest system uses a traditional trigram model
(Steetskamp 1995; henceforth tagger T, for Trigrams), based on context
statistics $P(t_i|t_{i-1},t_{i-2}$) and lexical statistics
$P(t_i|w_i)$ directly estimated from relative corpus frequencies. The
Viterbi algorithm is used to determine the most probable tag
sequence. Since this model has no facilities for handling unknown
words, a Memory-Based system (see below) is used to propose
distributions of potential tags for words not in the lexicon. 

The second system is the Transformation Based Learning system as described
by Brill (1994\footnote{Brill's system is available as a collection of
C programs and Perl scripts at 
{\tt ftp://ftp.cs.jhu.edu/pub/brill/Programs/ RULE\_BASED\_TAGGER\_V.1.14.tar.Z}}; henceforth
tagger R, for Rules). This system starts with a basic corpus
annotation (each word is tagged with its most likely tag) and then
searches through a space of transformation rules in order to reduce
the discrepancy between its current annotation and the correct one (in
our case 528 rules were learned). During tagging these rules are
applied in sequence to new text.
Of all the four systems, this one has access to the most information: 
contextual information (the words and tags in a window spanning
three positions before and after the focus word)
as well as lexical information (the existence of words formed
by suffix/prefix addition/deletion).
However, the actual use of this information is severely limited 
in that the individual information items can only be combined
according to the patterns laid down in the rule templates.

The third system uses Memory-Based Learning as described by Daelemans
{\em et al.} (1996; henceforth tagger M, for Memory). During the training phase, cases
containing information about the word, the context and the correct tag
are stored in memory. During tagging, the case most similar to that of
the focus word is retrieved from the memory, which is indexed on the
basis of the Information Gain of each feature, and the accompanying
tag is selected. The system used here has access to information about the focus
word and the two positions before and after, at least for known words. For
unknown words, the single position before and after, three suffix
letters, and information about capitalization and presence of a hyphen
or a digit are used. 

The fourth and final system is the MXPOST system as
described by Ratnaparkhi (1996\footnote{Ratnaparkhi's Java
implementation of this system is available at {\tt
ftp://ftp.cis.upenn.edu/ pub/adwait/jmx/}}; henceforth tagger E, for
Entropy). It uses a number of word and context features rather similar
to system M, and trains a Maximum Entropy model that assigns a
weighting parameter to each feature-value and combination of features
that is relevant to the estimation of the probability
$P(tag|features)$. A beam search is then used to find the highest
probability tag sequence. Both this system and Brill's system are used
with the default settings that are suggested in their documentation.

\section{The data}

The data we use for our experiment consists of the tagged
LOB corpus (Johansson 1986). The corpus comprises about one million
words, divided over 500 samples of 2000 words from 15 text types. Its
tagging, which was manually checked and corrected, is generally
accepted to be quite accurate.  Here we use a slight adaptation of the
tagset. The changes are mainly cosmetic, e.g. non-alphabetic
characters such as ``\$'' in tag names have been replaced. However, there
has also been some retokenization: genitive markers have been split
off and the negative marker ``n't'' has been reattached.  An example
sentence tagged with the resulting tagset is: 

\begin{table}[h]
\begin{center}
\begin{tabular}{l l p{3cm}}
The & ATI & singular or plural\\
& &         article\\
Lord & NPT & singular titular\\
& &          noun\\
Major & NPT & singular titular\\
& &           noun\\
extended & VBD & past tense of verb\\
an & AT & singular article\\
invitation & NN & singular common\\
& &  noun\\
to & IN & preposition\\
all & ABN & pre-quantifier\\
the & ATI & singular or plural\\
& & article\\
parliamentary & JJ & adjective\\
candidates & NNS & plural common\\
& & noun\\
. &  SPER & period\\
\end{tabular}
\end{center}
\end{table}

The tagset consists of 170 different tags (including ditto
tags\footnote{Ditto tags are used for the components of multi-token
units, e.g. if ``as well as'' is taken to be a coordination conjunction,
it is tagged ``as\_CC-1 well\_CC-2 as\_CC-3'', using three related but
different ditto tags.}) and has an average ambiguity of 2.69 tags per
wordform. The difficulty of the tagging task can be judged by the two
baseline measurements in Table~\ref{measurements} below, representing
a completely random choice from the potential tags for each token
(Random) and selection of the lexically most likely tag (LexProb).

For our experiment, we divide the corpus into three parts. The first
part, called Train, consists of 80\% of the data (931062 tokens),
constructed by taking the first eight utterances of every ten. This
part is used to train the individual taggers. The second part, Tune,
consists of 10\% of the data (every ninth utterance, 114479 tokens)
and is used to
select the best tagger parameters where applicable and
to develop the combination methods.
The third and final part, Test,
consists of the remaining 10\% (115101 tokens) and is used for the
final performance measurements of all taggers.
Both Tune and Test contain around 2.5\% new tokens (wrt Train)
and a further 0.2\% known tokens with new tags.

The data in Train (for
individual taggers) and Tune (for combination taggers) is to be the
only information used in tagger construction: all components of all
taggers (lexicon, context statistics, etc.) are to be entirely data
driven and no manual adjustments are to be done. The data in Test is
never to be inspected in detail but only used as a benchmark tagging
for quality measurement.%
\footnote{This implies that it is impossible to note if errors
counted against a tagger are in fact errors in the benchmark tagging.
We accept that we are measuring quality in relation to a specific
tagging rather than the linguistic truth (if such exists) and can only 
hope the tagged LOB corpus lives up to its reputation.} 

\section{Potential for improvement}

In order to see whether combination of the component taggers is likely
to lead to improvements of tagging quality, we first examine the
results of the individual taggers when applied to Tune.  
As far as we know this is also one of the first
rigorous measurements of the relative quality of different tagger
generators, using a single tagset and dataset and identical
circumstances.

The quality of the individual taggers (cf. Table~\ref{measurements}
below) certainly still leaves room for improvement, although
tagger E surprises us with an accuracy well above any results reported
so far and makes us less confident about the gain to be accomplished with
combination.  

\begin{table}[t]
\begin{center}
\begin{tabular}{|p{5cm}|r|}
\hline
All Taggers Correct & 92.49\\
Majority Correct ({\bf 3}-1,{\bf 2}-1-1) & 4.34\\
Correct Present, No Majority ({\bf 2}-2,{\bf 1}-1-1-1) & 1.37\\
Minority Correct ({\bf 1}-3,{\bf 1}-2-1) & 1.01\\
All Taggers Wrong & 0.78\\
\hline
\end{tabular}
\caption{Tagger agreement on Tune. The patterns between the brackets
give the distribution of {\bf correct}/incorrect tags over the systems.}
\label{agreement}
\end{center}
\end{table}

However, that there is room for improvement is not enough. 
As explained above, for combination to lead to improvement,
the component taggers must differ
in the errors that they make. That this is indeed the case can be seen
in Table~\ref{agreement}.
It shows that for 99.22\% of Tune, at least one tagger selects the
correct tag. However, it is unlikely that we will be able to identify
this tag in each case. We should rather aim for optimal selection in
those cases where the correct tag is not outvoted, which would ideally
lead to correct tagging of 98.21\% of the words (in Tune).  

\section{Simple Voting}

There are many ways in which the results of the component taggers can
be combined, selecting a single tag from the set proposed by these
taggers. In this and the following sections we examine a number of
them. The accuracy measurements for all of them are listed in
Table~\ref{measurements}.%
\footnote{For any tag X, {\em precision} measures which percentage of the
tokens tagged X by the tagger are also tagged X in the benchmark and 
{\em recall} measures which percentage of the tokens tagged X in the 
benchmark are also tagged X by the tagger.
When abstracting away from individual tags, precision and recall are
equal and measure how many tokens are tagged correctly; in this
case we also use the more generic term {\em accuracy}.}
The most straightforward selection method is
an n-way vote. Each tagger is allowed to vote for the tag of its
choice and the tag with the highest number of votes is selected.%
\footnote{In our experiment, a random selection
{}from among the winning tags is made whenever there is a tie.}

\begin{table}[t]
\begin{center}
\begin{tabular}{|l|r|r|}
\hline
 & {\bf Tune} & {\bf Test}\\
\hline
{\bf Baseline} & & \\
Random & 73.68 & 73.74\\
LexProb & 92.05 & 92.27\\
\hline
{\bf Single Tagger} & & \\
T & 95.94 & 96.08\\
R & 96.34 & 96.46\\
M & 96.76 & 96.95\\
E & 97.34 & 97.43\\
\hline
{\bf Simple Voting} & & \\
Majority & 97.53 & 97.63\\
TotPrecision & 97.72 & 97.80\\
TagPrecision & 97.55 & 97.68\\
Precision-Recall & 97.73 & 97.84\\
\hline
{\bf Pairwise Voting} &  & \\
TagPair & 97.99 & 97.92\\
\hline
{\bf Memory-Based} & &\\
Tags & 98.31 & 97.87\\
Tags+Word & 99.21 & 97.82\\
Tags+Context & 99.46 & 97.69\\
\hline
{\bf Decision trees} & & \\
Tags & 98.08 & 97.78\\
Tags+Word & -- & --\\
Tags+Context & 98.67 & 97.63\\
\hline
\end{tabular}
\caption{
Accuracy of individual taggers and combination methods.
}
\label{measurements}
\end{center}
\end{table}

The question is how large a vote we allow each tagger. The
most democratic option is to give each tagger one vote (Majority).
However, it appears more useful to give more weight to taggers which have
proved their quality. This can be general quality, e.g. each tagger
votes its overall precision (TotPrecision),
or quality in relation to
the current situation, e.g. each tagger votes its precision on the
suggested tag (TagPrecision). The information about each tagger's quality
is derived from an inspection of its results on Tune.  

But we have even more information on how well the taggers perform. We not only know
whether we should believe what they propose (precision) but also
know how often they fail to recognize the correct tag (recall). This
information can be used by forcing each tagger also to add to the
vote for tags suggested by the opposition, by an amount equal to 1
minus the recall on the opposing tag (Precision-Recall).

As it turns out, all voting systems outperform the best single tagger,
E.\footnote{Even the worst combinator, Majority, is significantly
better than E: using McNemar's chi-square, p=0.} Also, the best voting
system is the one in which the most specific information is used,
Precision-Recall. However, specific information is not always
superior, for TotPrecision scores higher than TagPrecision. This might
be explained by the fact that recall information is missing (for
overall performance this does not matter, since recall is equal to precision).  

\section{Pairwise Voting}

So far, we have only used information on the performance of individual
taggers. A next step is to examine them in pairs. We can investigate
all situations where one tagger suggests $T_1$ and the
other $T_2$ and estimate the probability that in this situation
the tag should actually be $T_x$, 
e.g. if E suggests DT and T suggests CS (which can happen if the token is
``that'') the probabilities for the appropriate tag are:

\begin{table}[h]
\begin{center}
\begin{tabular}{l l l}
CS & subordinating conjunction &0.3276\\
DT & determiner &0.6207\\
QL & quantifier &0.0172\\
WPR & wh-pronoun & 0.0345\\
\end{tabular}
\end{center}
\end{table}

When combining the taggers, every tagger pair is taken in turn and allowed
to vote (with the probability described above)
for each possible tag, i.e. not just the ones suggested by the
component taggers. If a tag pair $T_1$-$T_2$ has never been observed in Tune,
we fall back on information on the individual taggers, viz. the
probability of each tag $T_x$ given that the tagger suggested tag $T_i$.  

Note that with this method (and those in the next section) a tag suggested
by a minority (or even none) of the taggers still has a chance to win. In
principle, this could remove the restriction of gain only in 2-2 and
1-1-1-1 cases. In practice, the chance to beat a majority is very
slight indeed and we should not get our hopes up too high that this
should happen very often.  

When used on Test, the pairwise voting
strategy (TagPair) clearly outperforms the other voting
strategies,\footnote{It is significantly better than the runner-up
(Precision-Recall) with p=0.} but does not yet approach the level
where all tying majority votes are handled correctly (98.31\%).

\section{Stacked classifiers} 

{}From the measurements so far it appears that the use of more detailed
information leads to a better accuracy improvement. It ought therefore to be
advantageous to step away from the underlying mechanism of voting and
to model the situations observed in Tune more closely. The practice of
feeding the outputs of a number of classifiers as features for a next
learner is usually called {\em stacking} (Wolpert 1992). The
second stage can be provided with the first level outputs, and with
additional information, e.g. about the original input pattern. 

The first choice for this is to use a Memory-Based second level
learner. In the basic version (Tags), each case consists of the tags
suggested by the component taggers and the correct tag. In the more
advanced versions we also add information about the word in question
(Tags+Word) and the tags suggested by all taggers for the previous and
the next position (Tags+Context). For the first two the similarity
metric used during tagging is a straightforward overlap count; for the
third we need to use an Information Gain weighting (Daelemans {\em et al.} 1997).  

Surprisingly, none of the Memory-Based based methods
reaches the quality of TagPair.\footnote{Tags (Memory-Based) scores
significantly worse than TagPair (p=0.0274) and not significantly
better than Precision-Recall (p=0.2766).} The explanation for this can
be found when we examine the differences within the Memory-Based
general strategy: the more feature information is stored, the higher
the accuracy on Tune, but the lower the accuracy on Test. This 
is most likely an overtraining effect: Tune is probably too small to collect case
bases which can leverage the stacking effect convincingly, especially
since only 7.51\% of the second stage material shows disagreement
between the featured tags.

To examine if the overtraining
effects are specific to this particular second level classifier, we
also used the C5.0 system, a commercial version of the well-known program
C4.5 (Quinlan 1993) for the induction of decision trees, on the
same training material.%
\footnote{Tags+Word could not be handled by
C5.0 due to the huge number of feature values.} 
Because C5.0 prunes
the decision tree, the overfitting of training material (Tune) is less
than with Memory-Based learning, but the results on Test are also worse. 
We conjecture that pruning is not beneficial when the interesting cases
are very rare. To realise the benefits of stacking, either more data
is needed or a second stage classifier that is better suited to this
type of problem.


\begin{table}[t]
\begin{center}
\begin{tabular}{|l|r|p{2cm}|p{2cm}|}
\hline
 & Test & Increase vs Component Average & \% Reduction  Error Rate Best Component\\
\hline
T & 96.08 & - & -\\
R & 96.46 & - & -\\
M & 96.95 & - & -\\
MR & 97.03 & 96.70+0.33 & 2.6 (M) \\
RT & 97.11 & 96.27+0.84 & 18.4 (R) \\
MT & 97.26 & 96.52+0.74 & 10.2 (M) \\
E  & 97.43 & - &  - \\
MRT & 97.52 & 96.50+1.02 & 18.7 (M) \\
ME &  97.56 & 97.19+0.37 &  5.1 (E) \\
ER &  97.58 & 96.95+0.63 &  5.8 (E) \\
ET &  97.60 & 96.76+0.84 &  6.6 (E) \\
MER & 97.75 & 96.95+0.80 & 12.5 (E) \\
ERT & 97.79 & 96.66+1.13 & 14.0 (E) \\
MET & 97.86 & 96.82+1.04 & 16.7 (E) \\
MERT & 97.92 & 96.73+1.19 & 19.1 (E) \\
\hline
\end{tabular}
\caption{Correctness scores on Test for Pairwise Voting with all
tagger combinations}
\label{all-combis}
\end{center}
\end{table}

\section{The value of combination}

The relation between the accuracy of combinations (using TagPair) and
that of the individual taggers is shown in Table~\ref{all-combis}. The
most important observation is that every combination (significantly) 
outperforms the combination of any strict subset of its components.
Also of note is the improvement yielded by the best combination. 
The pairwise voting
system, using all four individual taggers, scores 97.92\% correct on
Test, a 19.1\% reduction in error rate over the best individual system,
viz. the Maximum Entropy tagger (97.43\%).

A major factor in the quality of the combination
results is obviously the quality of the best component: all
combinations with E score higher than those without E (although M, R
and T together are able to beat E alone%
\footnote{By a margin at the edge of significance: p=0.0608.}). After that, the
decisive factor appears to be the difference in language model: T is
generally a better combiner than M and R,\footnote{Although not
significantly better, e.g. the differences within the group ME/ER/ET
are not significant.} even though it has the lowest accuracy when
operating alone. 

A possible criticism of the proposed combination scheme
is the fact that for the most successful combination schemes, one has
to reserve a non-trivial portion 
(in the experiment 10\% of the total material)
of the annotated data to set the parameters for the combination. To see
whether this is in fact a good way to spend the extra data, we also
trained the two best individual systems (E and M, with exactly the
same settings as in the first experiments) on a concatenation of Train
and Tune, so that they had access to every piece of data that the
combination had seen. It turns out that the increase in the individual
taggers is quite limited when compared to combination. The more 
extensively trained E scored 97.51\% correct on Test (3.1\% error 
reduction) and M 97.07\% (3.9\% error reduction).

\section*{Conclusion}

Our experiment shows that, at least for the task at hand, combination
of several different systems allows us to raise the performance ceiling
for data driven systems. Obviously there is still room for a closer
examination of the differences between the combination methods,
e.g. the question whether Memory-Based combination would have performed
better if we had provided more training data than just Tune, and of
the remaining errors, e.g. the effects of inconsistency in the data
(cf. Ratnaparkhi 1996 on such effects in the Penn Treebank
corpus). Regardless of such closer investigation, we feel that our
results are encouraging enough to extend our investigation of
combination, starting with additional component taggers and selection
strategies, and going on to shifts to other tagsets and/or languages. But
the investigation need not be limited to wordclass tagging, for we
expect that there are many other NLP tasks where combination could
lead to worthwhile improvements.

\section*{Acknowledgements}

Our thanks go to the creators of the tagger generators used here for
making their systems available. 

\end{document}